\documentclass{article}
\usepackage{ijcai17}
\usepackage{times}
\usepackage{graphicx}
\usepackage{subfigure}
\usepackage{floatrow}
\usepackage{tikz}
\usepackage{amsmath}
\usepackage[tableposition=top]{caption}
\usepackage{multirow}
\usepackage{placeins}
\usepackage{float}
\floatstyle{plaintop}
\restylefloat{table}

\title{The Forgettable-Watcher Model for Video Question Answering}
\author{Hongyang Xue, Zhou Zhao, Deng Cai  \\
hyxue@outlook.com, zhaozhou@zju.edu.cn, dengcai@gmail.com}

\begin{document}

\maketitle

\begin{abstract}
A number of visual question answering approaches have been proposed recently, aiming at understanding the visual scenes by answering the natural language questions.  While the image question answering has drawn significant attention, video question answering is largely unexplored. 
 Video-QA is different from Image-QA since the information and the events are scattered among multiple frames. In order to better utilize the temporal structure of the videos and the phrasal structures of the answers,  we propose two mechanisms: the re-watching and the re-reading mechanisms and combine them into the forgettable-watcher model. Then we propose a TGIF-QA dataset for video question answering with the help of automatic question generation. Finally, we evaluate the models on our dataset. The experimental results show the effectiveness of our proposed models.
\end{abstract}

\section{Introduction}
To understand the visual scenes is one of the ultimate goals in computer vision. A lot of intermediate and low-level tasks, such as object detection, recognition, segmentation and tracking, have been studied towards this goal. One of the high-level tasks towards scene understanding is the visual question answering \cite{antol2015vqa}. This task aims at understanding the scenes by answering the questions about the visual data. It also has a wide application, from aiding the visually-impaired, analyzing surveillance data to domestic robots. 

The visual data we are facing everyday are mostly dynamic videos. 
However, most of the current visual question answering works focus only on images \cite{bigham2010vizwiz,geman2015visual,gao2015you,Yang_2016_CVPR,Noh_2016_CVPR,ma2016learning}. The images are static and contain far less information than the videos. The task of image-based question answering cannot fit into  real-world applications since it ignores the temporal coherence of the scenes. 

   Existing video-related question answering works usually combine additional information. The Movie-QA dataset \cite{MovieQA} contains multiple sources of information: plots, subtitles, video clips, scripts and DVS transcriptions. These extra information is hard to retrieve in the real world, making these datasets and approaches difficult to extend to general videos.
  
Unlike the previous works, we consider the more ubiquitous task of video question answering with only the visual data and the natural language questions. In our task, only the videos, the questions and the corresponding answer choices are presented. We first introduce a dataset collected on our own. To collect a dataset is not an easy task. In image-based question answering (Visual-QA) \cite{antol2015vqa}, most current collection methods require humans to generate the question-answer pairs \cite{antol2015vqa,malinowski2014multi}. This requires a significant amount of human labor. To make things worse, the video has a temporal dimension compared with the image, which implies that the labor of the human annotators is multiplied. To avoid the significant increase of human labor, our solution is to employ the question generation approaches \cite{heilman2010good} to generate question-answer pairs directly from the texts accompanying the videos. Now the collection becomes collecting videos with descriptions. This inspires us to utilize the existing video description datasets. The TGIF (Tumblr GIF) dataset \cite{Li_2016_CVPR} is a large-scale video description dataset. The groundtruth description provides us the necessary texts to produce the question-answer pairs.  Finally, we form our TGIF-QA dataset. Details will be described in the Dataset section.

\begin{figure}[!hbt]
	\begin{minipage}[t]{0.1\linewidth}
		\centering
		\includegraphics[scale=0.1]{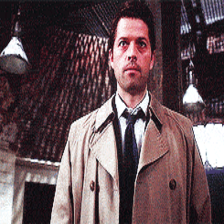}
		\label{fig:side:so}
	
	\end{minipage}%
	\begin{minipage}[t]{0.1\linewidth}
		\centering
		\includegraphics[scale=0.1]{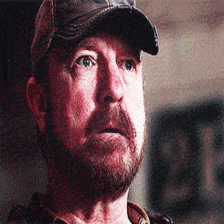}

		\label{fig:side:sp}
	\end{minipage}%
	\begin{minipage}[t]{0.1\linewidth}
		\centering
		\includegraphics[scale=0.1]{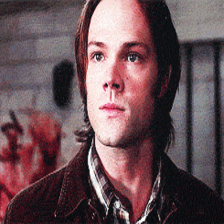}
		\label{fig:side:sq}
	
	\end{minipage}%
	\begin{minipage}[t]{0.1\linewidth}
		\centering
		\includegraphics[scale=0.1]{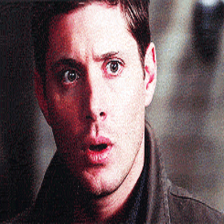}
		\label{fig:side:sqg}
		
	\end{minipage}%
	\begin{minipage}[t]{0.1\linewidth}
		\centering
		\includegraphics[scale=0.1]{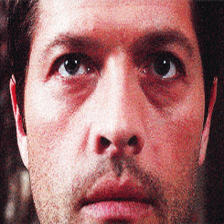}
		\label{fig:side:sqr}
		
	\end{minipage}
	\caption{Sample frames of a video clip. The question is ``How many men stare at each other while one of them rages?". This is a typical example where the information is scattered on multiple frames. The information along the temporal dimension needs to be accumulated to answer the question. A single frame or a small duration of the video is definitely not sufficient for the question.}
\end{figure}
Existing visual question answering approaches are not suitable for video question answering since the video question answering has the following features: First, the question may relate to an event which happens across multiple frames. The information must be gathered among the frames to answer the question. A typical case is the question related to the numbers (see in Fig~1.). The question asks about the number of men in the video. However, in the beginning, we cannot see the correct number of men. Only by watching the video frame by frame can we tell the answer is four. Same is the case in the top-right of Fig~2. Current existing visual-QA approaches cannot be applied as they only utilize the spatial information of a static image. Second, there may be a lot of redundancy in the video frames. In addition to these, our task faces another challenge: the candidate answers are mostly phrases. To tackle these problems, we propose two models: the re-watcher and the re-reader. The re-watcher model meticulously processes the video frames. This model allows to gather information from relevant frames. Then the information is recurrently accumulated as the question is read. The re-reader model can handle phrasal answers and concentrate on the important words in the answers. Then we combine these two models into the forgettable-watcher model.

Our contribution can be summarized into two aspects: First we introduce a Video-QA dataset TGIF-QA. Second we propose the models which can employ the temporal structure of the videos and the phrasal structure of the candidate answers. We also extend the VQA model \cite{antol2015vqa} as a baseline method.
We evaluate our models on our proposed TGIF-QA dataset.  The experimental results show the effectiveness of our proposed models.

\begin{figure*}[!hbt]
	
	\begin{minipage}[t]{0.11\textwidth}
		\centering
		\includegraphics[width=1.6cm]{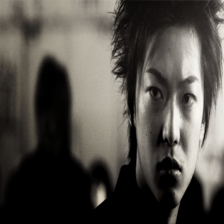}
		\label{fig:side:a}
	\end{minipage}%
	\begin{minipage}[t]{0.11\textwidth}
		\centering
		\includegraphics[width=1.6cm]{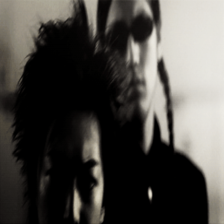}
		\label{fig:side:b}
	\end{minipage}%
	\begin{minipage}[t]{0.11\textwidth}
		\centering
		\includegraphics[width=1.6cm]{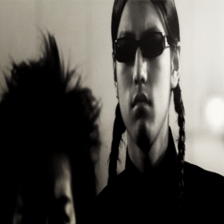}
		\label{fig:side:c}
	\end{minipage}
	\begin{minipage}[t]{0.11\textwidth}
		\centering
		\includegraphics[width=1.6cm]{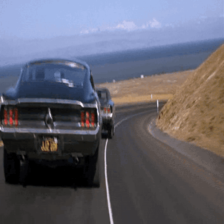}
		\label{fig:side:a1}
	\end{minipage}%
	\begin{minipage}[t]{0.11\textwidth}
		\centering
		\includegraphics[width=1.6cm]{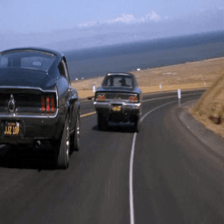}
		\label{fig:side:b1}
	\end{minipage}%
	\begin{minipage}[t]{0.11\textwidth}
		\centering
		\includegraphics[width=1.6cm]{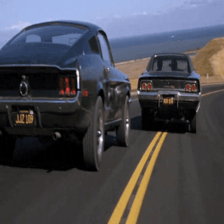}
		\label{fig:side:c1}
	\end{minipage}
	
	\begin{minipage}[t]{6cm}
		
		{\scriptsize D: a man is glaring, and someone\\ with sunglasses appears.\\ Q: Who is glaring?\\
			GT: a man\hspace{0.5cm}\hspace{0.5cm} A1: a jumper\\
			A2: a spectator\hspace{0.52cm} A3: a husband\\
			A4: a bowler\hspace{0.67cm} A5: a artist\\
			A6: a guitarist\hspace{0.5cm} A7: a dog}
	\end{minipage}%
	\begin{minipage}[t]{7cm}
		{\scriptsize D: two cars is chasing each other along a highway \\Q: How many cars is chasing each other along a highway?\\
			GT: two cars\hspace{0.5cm}  A1: eight cars\\
			A2: one cars\hspace{0.5cm} A3: five cars\\
			A4: four cars\hspace{0.43cm} A5: seven cars\\
			A6: six cars\hspace{0.6cm} A7: three cars}
	\end{minipage}
	
	\vspace{0.5cm}
	
	\begin{minipage}[t]{0.11\textwidth}
		\centering
		\includegraphics[width=1.6cm]{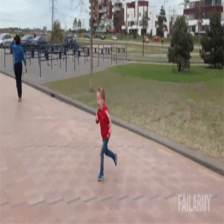}
		\label{fig:side:a2}
	\end{minipage}%
	\begin{minipage}[t]{0.11\textwidth}
		\centering
		\includegraphics[width=1.6cm]{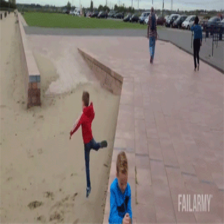}
		\label{fig:side:b2}
	\end{minipage}%
	\begin{minipage}[t]{0.11\textwidth}
		\centering
		\includegraphics[width=1.6cm]{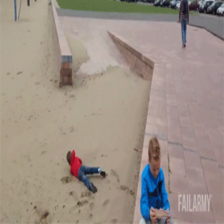}
		\label{fig:side:c2}
	\end{minipage}
	\begin{minipage}[t]{0.11\textwidth}
		\centering
		\includegraphics[width=1.6cm]{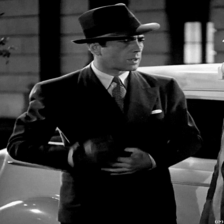}
		\label{fig:side:a3}
	\end{minipage}%
	\begin{minipage}[t]{0.11\textwidth}
		\centering
		\includegraphics[width=1.6cm]{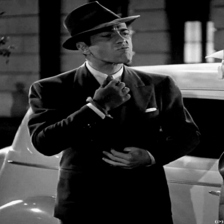}
		\label{fig:side:b3}
	\end{minipage}%
	\begin{minipage}[t]{0.11\textwidth}
		\centering
		\includegraphics[width=1.6cm]{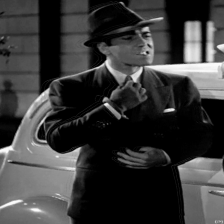}
		\label{fig:side:c3}
	\end{minipage}
	\begin{minipage}[t]{7cm}
		{\scriptsize D: a little boy runs across the pavement and goes to jump \\into the sand, but his leg gets caught and he falls instead \\Q: Whose leg gets caught?\\
			GT: a little boy 's leg\hspace{0.3cm}  A1: a little spectator 's leg\\
			A2: a little pilot 's leg\hspace{0.2cm} A3: a little performer 's leg\\
			A4: a little alien 's leg\hspace{0.2cm} A5: a little model 's leg\\
			A6: a little soldier 's leg\hspace{0.1cm} A7: a little biker 's leg}
	\end{minipage}%
	\begin{minipage}[t]{6cm}
		{\scriptsize D: a man is adjusting his tie and strokes his face \\Q: What does a man in a hat adjust?\\
			GT: his tie \hspace{1.2cm}  A1: his scar\\
			A2: his mannequin\hspace{0.4cm} A3: his trunk\\
			A4: his watch\hspace{0.85cm} A5: his caps\\
			A6: his pattern\hspace{0.73cm} A7: his pocket}
	\end{minipage}
	
	\caption{Sample videos and question  pairs in our TGIF-QA dataset. Each video is provided with its original description, a question and 8 candidate answers. The ground-truth answer is the one closest to the scene and the question. D indicates the description. Q indicates the question. GT is the ground-truth answer. A1-A7 are the wrong alternatives.}
\end{figure*}
\section{Dataset}

In this section, we introduce our dataset for Video Question Answering. We convert the existing video description dataset for question answering. The TGIF dataset \cite{Li_2016_CVPR} is collected by Li \textit{et al.} from Tumblr. GIFs are almost identical to small video clips for the short duration. Li \textit{et al.} cleaned up the original data and ruled out the GIFs with catoon, static and textual content. The animated GIFs were later annotated using crowd-sourcing service. 
The TGIF dataset contains 102,068 GIFs and 287,933 descriptions in total where each GIF corresponds to several descriptions. Each description consists of one or two sentences.

\subsection{TGIF-QA}
In order to generate the question-answer pairs from the descriptions, we employ the state-of-the-art question generation approach  \cite{heilman2010good}. We focus on the questions of types \textbf{What}/\textbf{When}/\textbf{Who}/\textbf{Whose}/\textbf{Where} and \textbf{How Many}. Our question answering task is of the multiple-choice type, and the generated data only contains question and ground-truth answer pairs. So we need to generate wrong alternatives for each question. We provide each question with 8 candidate answers. We describe how we generate the alternative answers for each kind of questions in the following subsections.

\subsubsection{How Many}
The \textbf{How Many} question relates to counting some objects in the video. In order to generate reasonable alternatives, we first collect all the \textbf{How Many} questions in our dataset and gather the numbers in their answers. All the Arabic numerals are converted to English words representations. After eliminating the numbers of low occurrence frequency, we find that most answers contain numbers from one to eight. We discard the questions whose answers exceed eight and replace the ground-truth numbers with \textbf{[one, two, three, four, five, six, seven, eight]} to generate the 8 candidate answers. One typical example is shown in the top-right of Fig~1.

\subsubsection{Who}
The questions starting with \textbf{Who} usually relate to humans. We collect the words in the answers from all the \textbf{Who} questions. Then we filter the words to obtain all the nouns. After that, we filter out all the abstract and low-frequency nouns to form an entity list. The entity word in the ground-truth answers is selected and replaced with random samples from the entity list to generate 8 alternatives. An example is provided in the top-left of Fig~1.

\subsubsection{Whose}
The \textbf{Whose} questions relate to the facts about belongings. There are two ways to represent the belongings. One is through the possessive pronoun like ``my", ``your", ``his", etc. The other is to use the possessive case of nouns such as ``man's", ``girl's", ``cats'", etc. For the former kind of possessive pronouns, we replace the pronouns with random samples from the possessive pronoun list to generate the alternatives. For the latter one we replace the nouns just the same way we do for the \textbf{Who} questions.

\subsubsection{Other Questions}
For the rest types of questions, we just replace the nouns in the answer phrases to generate candidate choices. 

\subsection{Dataset Split}
We abandon the videos whose descriptions generate no questions or the questions have been discarded in the previous processing. In the end our dataset contains 117,854 videos and 287,763 question-candidates pairs.
We split our TGIF-QA dataset into three parts for training, validation and testing. The training dataset contains 230,689 question-answers pairs from 94,425 videos. The validation dataset contains 24,696 pairs from 10,096 videos and the testing dataset has 32,378 pairs from 13,333 videos.

\section{Method}
\subsection{Task description}
Multiple-Choice Video-QA is to select an answer $\mathbf{\tilde{a}}$ given a question $\mathbf{q}$, video information $\mathbf{v}$ and candidate choices (alternatives) $\mathbf{a} = \{\mathbf{a}_1, \cdots, \mathbf{a}_8\}$. A video is a sequence of image frames $\mathbf{v} = \{f_1, f_2, \cdots\}$. A question is a sequence of natural language words $\mathbf{q} = \{q_1, q_2, \cdots\}$. Each alternative of the candidate answers is also a sequence of natural language words $\mathbf{a}_i = \{a^1_i, a^2_i, \cdots \}$. We formulate the Video-QA task as selecting the best answer among the alternatives. In the other word, we define a loss function $L(\mathbf{v}, \mathbf{q}, \mathbf{a}_i)$. We regard the QA problem as a classification problem and the best answer is selected when it achieves minimal loss $\mathbf{\tilde{a}} = \arg\max_{\mathbf{a}_i}L(\mathbf{v},\mathbf{q},\mathbf{a}_i)$.

\subsection{Model}

\begin{figure*}[!hbt]
	\begin{minipage}[t]{0.3\linewidth}
		\centering
		\includegraphics[scale=0.19]{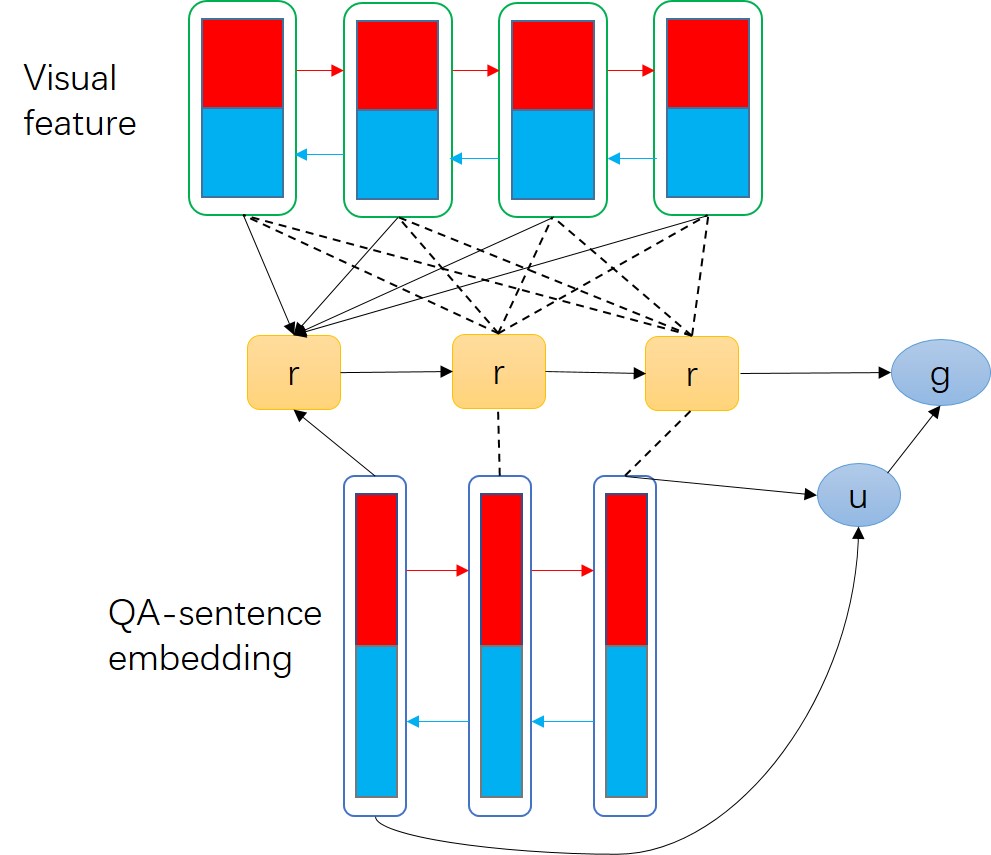}
		\label{fig:side:o}
	
	\end{minipage}%
	\begin{minipage}[t]{0.3\linewidth}
		\centering
		\includegraphics[scale=0.19]{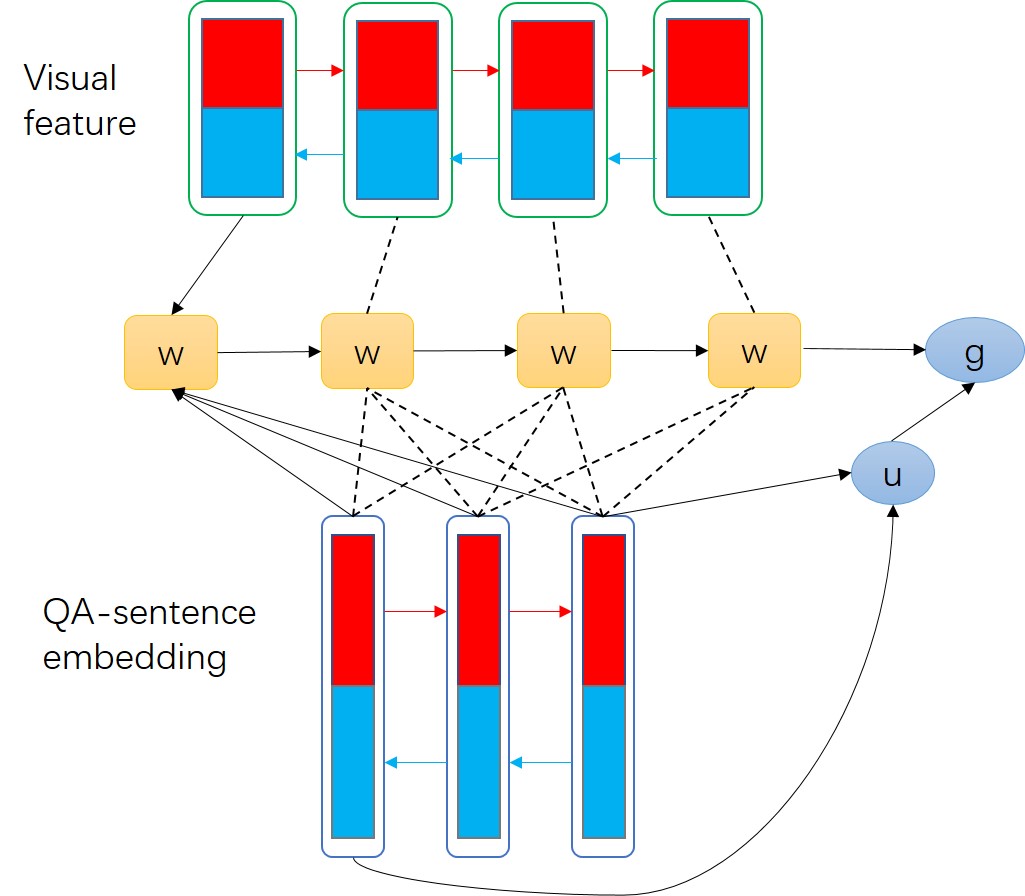}

		\label{fig:side:p}
	\end{minipage}%
	\begin{minipage}[t]{0.3\linewidth}
		\centering
		\includegraphics[scale=0.19]{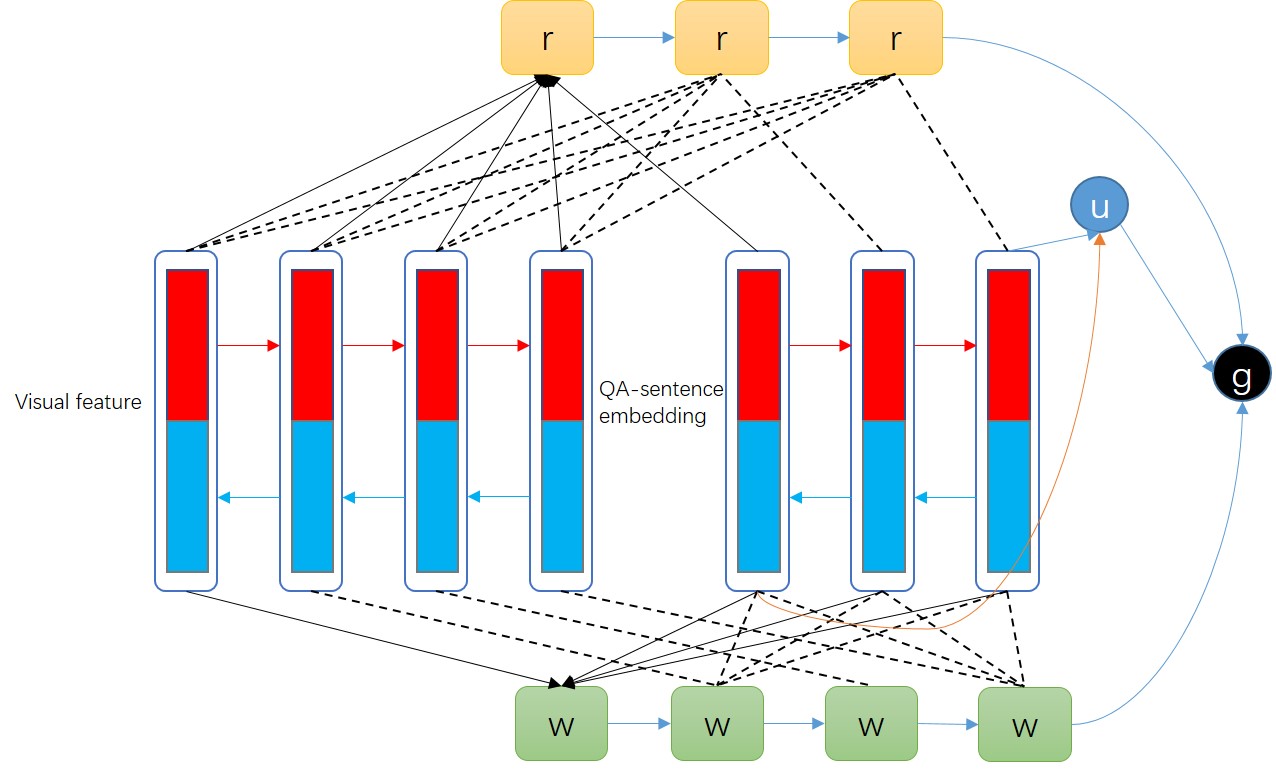}
		\label{fig:side:q}
	
	\end{minipage}
	\caption{Our proposed three models. From left to right are the re-watcher, re-reader and forgettable-watcher models respectively.}
\end{figure*}
The major difficulty of Video-QA compared with Image-QA is that the video has a lot of frames. The information is scattered among the frames and an event can last across the frames. To answer a question, we must find the most informative frame or combine the information of several frames.

In the following we propose three models: the re-watcher, the re-reader and the forgettable-watcher. The re-watcher model processes the question sentence word sequence meticulously. The re-reader model fully utilizes the information of the video frames along the temporal dimension. And the forgettable-watcher combines both of them.

We denote $\mathbf{qa}_i$ as the concatenation of question $\mathbf{q}$ and answer $\mathbf{a}_i$ after word embedding. All our models will take the question and one alternative as a whole sentence which we call the QA-sentence. The QA-sentence and the visual feature sequence are then put into our models to produce a score for the alternative answer in the sentence. In the following sections, we will denote the QA-sentence as $c$ when it does not introduce confusions.
\subsubsection{Re-Watcher}
Our model first encodes the video features and QA features with two separate bi-directional single layer LSTMs \cite{graves2012neural}. The LSTMs contain an embedding layer which maps the video features and textual features into a joint feature space. We denote the outputs of the forward and backward LSTMs as $y^f(t)$ and $y^b(t)$. The encoding $u$ of a QA-sentence of length $|c|$ is formed by the concatenation of final outputs of the forward and backward LSTMs, $u = [y_c^f(|c|),y_c^b(1)]$.

For the video frames, the encoding output for each frame at time $t$ is 
$y_v(t) = [y^f(t),y^b(t)]$. The representation $r(i)$ of the videos for each QA-sentence token $i$ is formed by a weighted sum of these output vectors (similar to the attention mechanism in Image-QA \cite{Yang_2016_CVPR}) and the previous representation $r(i-1)$:
\begin{equation*}
\begin{aligned}
m(i,t) &= \tanh(W_{vm}y_v(t)+W_{rm}r(i-1)+W_{cm}y_c(i))\\
s(i,t) &\propto \exp(W^T_{ms}m(i,t))\\
r(i) &= y^T_ds(i) + \tanh(W_{rr}r(i-1))\\
1&\le i\le |c|
\end{aligned}
\end{equation*}

The mechanism of the re-watcher model is that every word will combine with the whole video sequence to generate a state. Then the states of the word sequence is accumulated to generate a combined feature (see Fig~2. left). This model mimics a person who has a bad memory for the video he watches. Every time he reads a word of the QA-sentence, he goes back to watch the whole video to make out what the words in the sentence are about. During this procedure, he selects the information most related to the QA-sentence token from the video and then recurrently accumulate the information as the whole QA-sentence is read. Finally a joint video and QA-sentence representation is formed for producing a score. The score measures how much the question-answer pair (QA-sentence) matches with the video:
\begin{equation*}
g^{ReW} = \mathbf{FC}(\tanh(W_{rg}r(|c|)+W_{cg}u))
\end{equation*}
where $\mathbf{FC}$ represents three fully-connected layers. The activation of the former two layers is ReLU and the last layer is without activation.
Since each question has 8 alternatives, each question relates to 8 QA-sentences. For a question and 8 alternatives, we generate 8 such scores and they fill up a 8-dimensional score vector:
\begin{equation*}
g = [g^{ReW}_1,\cdots,g^{ReW}_8]
\end{equation*}
The score vector is then put through the softmax layer.

\subsubsection{Re-Reader}
The re-watcher model mimics a person who continuously re-watches the video as he reads the QA-sentence. Video features related with the QA-sentence are accumulated. The re-reader model is designed from the opposite view (see Fig~2. middle).

This model mimics a person who cannot well remember the whole question. Every time he watches a frame, he retrospects on the whole question. We denote the encoding output of the QA-sentence at token $i$ as $y_c(i) = [y^f_c(i), y^b_c(i)]$. The representation $w(t)$ of the video frames at time $t$ is computed from the weighted sum of the QA-sentence encoding outputs and the previous representation $w(t-1)$:
\begin{equation*}
\begin{aligned}
m(t,i) &= \tanh(W_{cm}y_c(i)+W_{wm}w(t-1)+W_{vm}y_v(t))\\
s(t,i) &\propto \exp(W^T_{ms}m(t,i))\\
w(t) &= y^T_cs(t) + \tanh(W_{ww}w(t-1))\\
1&\le t\le N
\end{aligned}
\end{equation*}
where $N$ is the number of frames. The score of the QA-sentence is:
\begin{equation*}
g^{ReR} = \mathbf{FC}(\tanh(W_{wg}w(N)+W_{cg}u))
\end{equation*}
\subsubsection{Forgettable-Watcher}
We combine the re-watcher and the re-reader models into the forgettable-watcher model (see Fig~2 right). This model meticulously combines the visual features and the texual features. The whole video is re-watched when a word is read and the whole sentence is re-read while a frame is being watched. Then the representations are combined to produce the score:
\begin{equation*}
g^F = \mathbf{FC}(\tanh(W_{rg}r(|c|) + W_{wg}w(N)+W_{cg}u))
\end{equation*}

\subsubsection{Baseline method}

In order to show the effectiveness of the re-reading and the re-watching mechanisms of our models. We also employ a straightforward model (see Fig~3). This model is extended from the VQA model \cite{antol2015vqa}. The VQA model is designed for question answering given only a single image. We extend the model for our task by directly encoding the video frames and the QA-sentences with two separate bidirectional LSTMs. The final encoding outputs of both bidirectional LSTMs are then combined to produce the score.
\begin{figure}[!hbt]

		\includegraphics[scale=0.19]{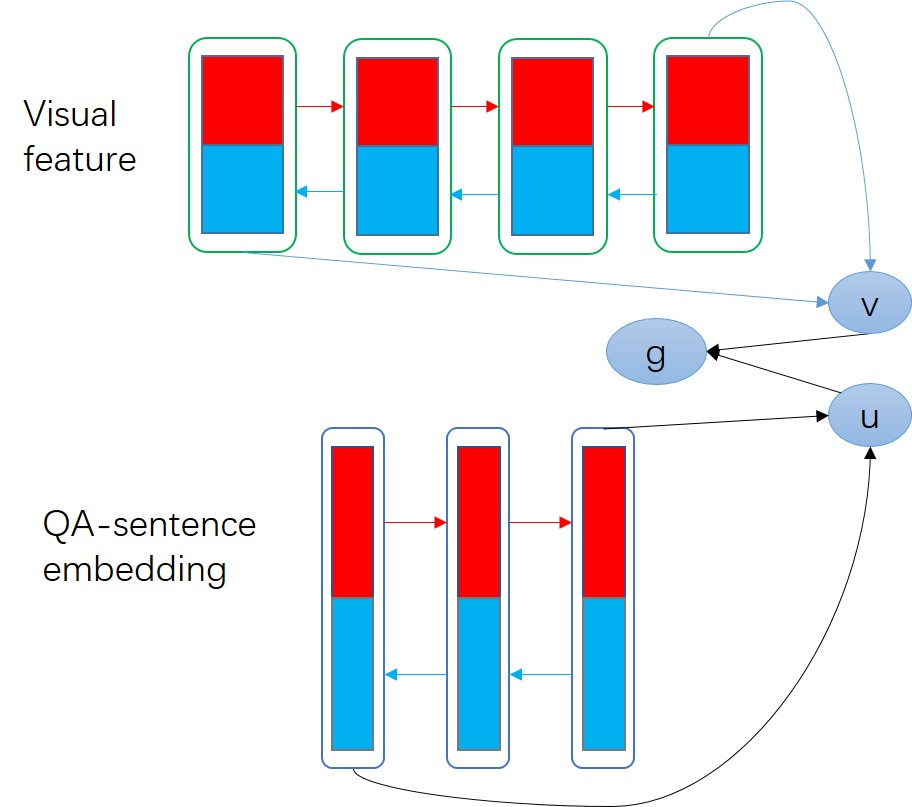}
\label{fig3}
\caption{The straightforward model without re-reading or re-watching mechanisms.}
\end{figure}
\section{Experiments and Results}
We evaluate all the methods on our TGIF-QA dataset described in the Dataset section. 
\subsection{Data Preprocessing}
We sample all the videos to have the same number of frames with the purpose of reducing the redundancy. If the video does not own enough frames for sampling, its last frame is repeated. We extract the visual features of each frame with VGGNet \cite{Simonyan14c}. The 4096-dimensional feature vector of the first fully-connected layer is taken as the visual feature. For the questions and answers, the Word2Vec \cite{mikolov} trained on GoogleNews is employed to embed each word as a 300-dimensional real vector. Then we concatenate each question with its 8 alternatives to generate 8 candidate QA-sentences.

\subsection{Implementation Details}
\subsubsection{Optimization}
We train all our models using the Adam optimizer \cite{kingma2014adam} to minimize the loss. The initial learning rate is set to 0.002. The exponential decay rates for the first and the second moment estimates are set to 0.1 and 0.001 respectively. The batch size is set to 100. A gradient clipping scheme is applied to clip gradient norm within 10. An early stopping strategy is utilized to stop training when the validation accuracy no longer improves.
\subsubsection{Model Details}
The visual features and the word embeddings are encoded by two separate bidirectional LSTMs to dimensionality 2048 and 1024 respectively. Then they are mapped to a joint feature space of dimensionality 1024. The re-watcher (re-reader) component keeps a memory of size 512 and outputs the final combined feature of dimensionality 512. Finally the combined feature is put into three fully-connected layers of size $(512, 256, 128)$. We evaluate our Video-QA task using classification accuracy.

\subsubsection{Evaluation Metrics}
Since our task is the multiple-choice question answering, we employ the classification accuracy to evaluate our models. However, there are a few cases where both the two choices can answer the question. This motivates us to also apply the WUPS measure \cite{malinowski2014multi} with $0.0$ and $0.9$ as the threshold values like the open-ended tasks.

\begin{table*}[!hbt]
	\caption{Video-QA results. We evaluate the baseline method in the first row(Straightforward). Our proposed three models are reported in the subsequent rows. Accuracy denotes the classification accuracy when the alternatives for each question are regarded as classes. WUPS $0.0$ and WUPS $0.9$ are the WUPS scores with $0.0$ and $0.9$ as threshold respectively.}
\begin{tabular}{|c|c|c|c|}
	\hline
	\multirow{2}{*}{Methods} & 
	\multicolumn{3}{|c|}{Evaluation} \\
	\cline{2-4} & Accuracy & WUPS@0.0 (\%) & WUPS@0.9 (\%)\\
	\hline
	\hline
	Straightforward & $0.8253$ & $93.24$ & $87.27$ \\
	Re-Watcher & $0.8663$ & $95.66$ & $91.28$ \\
	Re-Reader & $0.8592$ & $95.22$ & $90.75$ \\
	Forgettable & $\mathbf{0.8733}$ & $95.88$ & $\mathbf{92.56}$ \\
	\hline
\end{tabular}

\end{table*}
\makeatletter
\newcommand*{\textoverline}[1]{$\overline{\hbox{#1}}\m@th$}
\makeatother
\begin{table*}[!htb]
	\caption{This table shows the accuracy on different kinds of questions. We can see from the results that the \textbf{How many} questions are the most difficult. On the contrary, the questions asking \textbf{Where} and \textbf{When} are simpler. Doubting the alternatives for these two questions are not of high-quality, we also report the results on all the questions besides these two. It is denoted by \textoverline{Where and When}. }
\begin{tabular}{|c|c|c|c|c|c|c|c|}
	\hline
	\multirow{2}{*}{Methods} & 
	\multicolumn{7}{|c|}{Accuracy}  \\
	\cline{2-8} &  How many & Who & Whose & What & Where & When & \tiny{\textoverline{Where and When}}\\
	\hline
	\hline
	Straightforward & $0.8203$ & $0.8399$ & $0.7826$ & $0.8080$ & $0.9303$ & $0.9677$ & $0.8228$\\
	Re-Watcher  & $0.7808$ & $0.8931$ & $0.8447$ & $0.8461$ & $0.9467$ & $0.9596$ & $0.8646$\\
	Re-Reader  & $0.8069$ & $0.8843$ & $0.8323$ & $0.8373$ & $0.9467$ & $0.9704$ & $0.8572$\\
	Forgettable & $0.8233$ & $0.8984$ & $0.8634$ & $0.8515$ & $0.9468$ & $0.9730$ & $0.8715$\\
	\hline
\end{tabular}

\end{table*}

\begin{figure*}[!hbt]
\begin{minipage}[t]{0.4\textwidth}
	\begin{minipage}[t]{0.15\textwidth}
		\centering
		\includegraphics[width=1cm]{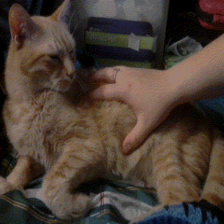}
		\label{fig:side:a}
	\end{minipage}%
	\begin{minipage}[t]{0.15\textwidth}
		\centering
		\includegraphics[width=1cm]{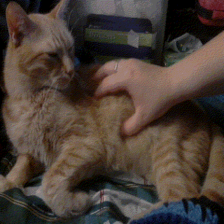}
		\label{fig:side:b}
	\end{minipage}%
	\begin{minipage}[t]{0.15\textwidth}
		\centering
		\includegraphics[width=1cm]{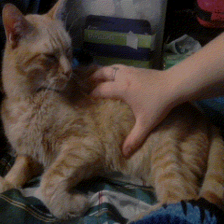}
		\label{fig:side:c}
	\end{minipage}%
		\begin{minipage}[t]{0.15\textwidth}
			\centering
			\includegraphics[width=1cm]{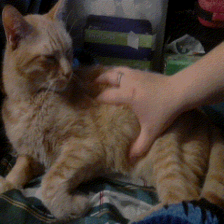}
			\label{fig:side:a1}
		\end{minipage}%
		\begin{minipage}[t]{0.15\textwidth}
			\centering
			\includegraphics[width=1cm]{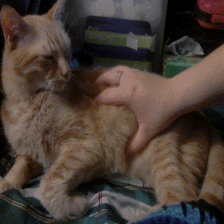}
			\label{fig:side:b1}
		\end{minipage}%
		\begin{minipage}[t]{0.15\textwidth}
			\centering
			\includegraphics[width=1cm]{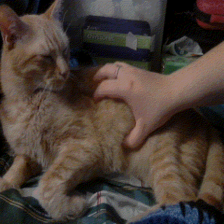}
			\label{fig:side:c1}
		\end{minipage}
\end{minipage}%
\begin{minipage}[t]{0.4\textwidth}
				\begin{minipage}[t]{0.15\textwidth}
					\centering
					\includegraphics[width=1cm]{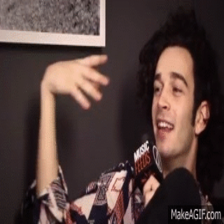}
					\label{fig:side:a2}
				\end{minipage}%
				\begin{minipage}[t]{0.15\textwidth}
					\centering
					\includegraphics[width=1cm]{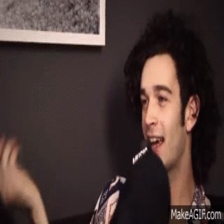}
					\label{fig:side:b2}
				\end{minipage}%
				\begin{minipage}[t]{0.15\textwidth}
					\centering
					\includegraphics[width=1cm]{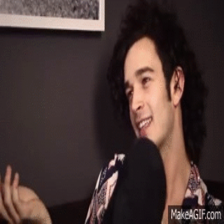}
					\label{fig:side:c2}
				\end{minipage}%
					\begin{minipage}[t]{0.15\textwidth}
						\centering
						\includegraphics[width=1cm]{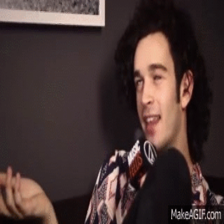}
						\label{fig:side:a3}
					\end{minipage}%
					\begin{minipage}[t]{0.15\textwidth}
						\centering
						\includegraphics[width=1cm]{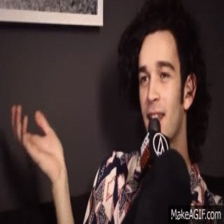}
						\label{fig:side:b3}
					\end{minipage}%
					\begin{minipage}[t]{0.15\textwidth}
						\centering
						\includegraphics[width=1cm]{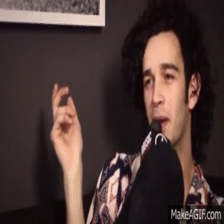}
						\label{fig:side:c3}
					\end{minipage}
				\end{minipage}
				
				\begin{minipage}[t]{0.4\textwidth}
						\begin{minipage}[t]{0.15\textwidth}
							\centering
							\includegraphics[width=1cm]{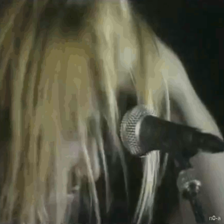}
							\label{fig:side:a2d}
						\end{minipage}%
						\begin{minipage}[t]{0.15\textwidth}
							\centering
							\includegraphics[width=1cm]{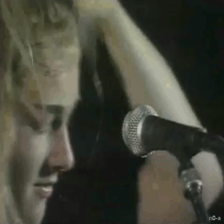}
							\label{fig:side:b2d}
						\end{minipage}%
						\begin{minipage}[t]{0.15\textwidth}
							\centering
							\includegraphics[width=1cm]{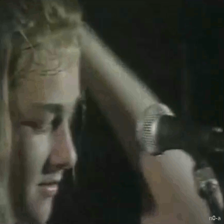}
							\label{fig:side:c2d}
						\end{minipage}%
						\begin{minipage}[t]{0.15\textwidth}
							\centering
							\includegraphics[width=1cm]{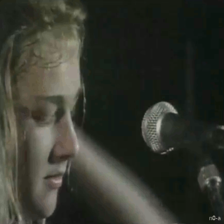}
							\label{fig:side:a3d}
						\end{minipage}%
						\begin{minipage}[t]{0.15\textwidth}
							\centering
							\includegraphics[width=1cm]{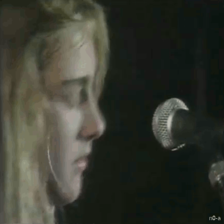}
							\label{fig:side:b3d}
						\end{minipage}%
						\begin{minipage}[t]{0.15\textwidth}
							\centering
							\includegraphics[width=1cm]{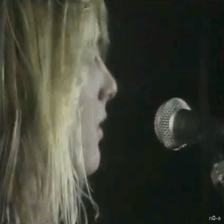}
							\label{fig:side:c3d}
						\end{minipage}
						\end{minipage}%
						\begin{minipage}[t]{0.4\textwidth}
							\begin{minipage}[t]{0.15\textwidth}
								\centering
								\includegraphics[width=1cm]{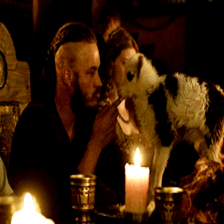}
								\label{fig:side:a2d}
							\end{minipage}%
							\begin{minipage}[t]{0.15\textwidth}
								\centering
								\includegraphics[width=1cm]{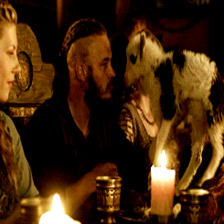}
								\label{fig:side:b2d}
							\end{minipage}%
							\begin{minipage}[t]{0.15\textwidth}
								\centering
								\includegraphics[width=1cm]{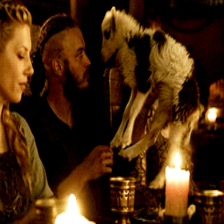}
								\label{fig:side:c2d}
							\end{minipage}%
							\begin{minipage}[t]{0.15\textwidth}
								\centering
								\includegraphics[width=1cm]{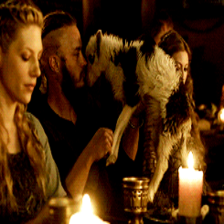}
								\label{fig:side:a3d}
							\end{minipage}%
							\begin{minipage}[t]{0.15\textwidth}
								\centering
								\includegraphics[width=1cm]{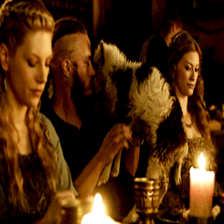}
								\label{fig:side:b3d}
							\end{minipage}%
							\begin{minipage}[t]{0.15\textwidth}
								\centering
								\includegraphics[width=1cm]{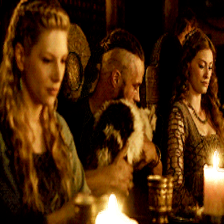}
								\label{fig:side:c3d}
							\end{minipage}
						\end{minipage}
%
					\begin{minipage}[t]{0.9\textwidth}
					\scriptsize	Q: What is a pet cat looking at? Pred: at the cat. GT: at the hand. \hspace{3cm} Q: Who talks. Pred: a woman. GT: a guy.\\ Q: Who is standing in front of a microphone? Pred: a young woman. GT: a young woman. \\Q: What is the man holding a lamb on? Pred: on the table. GT: on the table.
						
						\end{minipage}
						\caption{We exhibit four Video-QA examples. The two examples in the bottom row are our good examples. The two in the top row are failure cases.}
\end{figure*}
\subsection{Results}

\subsubsection{Baseline method}
The baseline method is an extension of the VQA model \cite{antol2015vqa} without our re-watching or re-reading mechanisms. We name it the straightforward model and its result is shown in the Straightforward section of table 1. We can see that the straightforward model performs much worse than the other three models.
\subsubsection{Our methods}
The forgettable-watcher model outperforms the other two models since it jointly employs the re-watching and the re-reading mechanisms. On the other hand, the re-reader model performs worse than the re-watcher model. This implies that the re-watching mechanism is more important.

\subsubsection{Results on different questions}
We also report the accuracy of the models on different types of questions. The result is shown in table~2. All the methods perform better on the questions asking about ``Where" and ``When" than others. This may be attributed to two reasons: First, the ``Where" and ``When" questions are easier to answer because these two questions usually relate to the video big scene. In most cases, a single frame may be enough to answer the questions. The other is that the candidate alternatives produced in the dataset generation may be too simple to discriminate. The dataset generation method is less effective in producing good alternatives for these two questions while it can produce high-quality alternatives for the other types of questions. We also report the results of the questions besides these two in table~2.

Finally, we exhibit some typical Video-QA results in Fig~5.

\section{Related Work}

\subsection{Image-QA}
Image-based visual question answering has attracted a significant research interest recently. \cite{bigham2010vizwiz,geman2015visual,antol2015vqa,gao2015you,Yang_2016_CVPR,Noh_2016_CVPR}. The goal of Image-QA is to answer questions given only one image without additional information. Image-QA tasks can be categorized into two types according to the ways answers are generated. The first type is called the Open-Ended question answering \cite{ren2015exploring,malinowski2014multi} where answers are produced given only the questions. As the answers generated by the algorithms are usually not the exact words people expect, measures such as the WUPS 0.9 and WUPS 0.0 based on the Wu-Palmer (WUPS) similarity \cite{malinowski2014multi} are employed to measure the answer accuracy. The other type is called the Multiple-Choice question answering \cite{Zhu_2016_CVPR} where both the question and several candidate answers are presented. To predict the answer is to pick the correct one among the candidates. It is observed that the approaches for the Open-Ended question answering usually cannot produce high-quality long answers \cite{Zhu_2016_CVPR}. And most Open-Ended question answering approaches only focus on one-word answers \cite{ren2015exploring,malinowski2014multi}. As a result, we consider the Multiple-Choice question answering type for our video question answering task. Our candidate answer choices are mainly phrases rather than single words.

A lot of efforts have been made tackling the Image-QA problem. Some of them have collected their own datasets. Malinowski et al \cite{malinowski2014multi} collected their dataset with the human annotations. Their dataset only focuses on basic colors, numbers and objects. Antol et al. \cite{antol2015vqa} manually collected a large-scale free-form Image-QA dataset. Gao et al. \cite{gao2015you} also manually collected the FM-IQA (Freestyle Multilingual Image Question Answering) dataset with the help of the Amazon Mechanical Turk platform. Most of these methods require a large amount of human labor for collecting data. In contrast, we propose to automatically convert existing video description dataset \cite{Li_2016_CVPR} into question answering dataset.

\subsection{Question Generation}

Automatic question generation is an open research topic in natural language processing. It is originally proposed for educational purpose \cite{gates2008automatically}. In our situation, we need the generated questions to be as diverse as possible so that it can well match the property of questions generated by human annotators. Among the question generation approaches \cite{rus2009question,gates2008automatically,heilman2010good}, we employ the method from Heilman and Smith \cite{heilman2010good} to generate our video-QA pairs from video description datasets. Their approach generates questions in open domains. Similar idea has been utilized by Ren et al. \cite{ren2015exploring} to turn image description datasets into Image-QA datasets. They generate only four types of questions: objects, numbers, color and locations. Their answers are mostly single words. On the contrary, we generate a large amount of open-domain questions where the corresponding answers are basically phrases.

\subsection{Video-QA}

Video-based question answering is a largely unexplored problem compared with Image-QA. Previous work usually combine other text information. Tapaswi et al. \cite{MovieQA} combine videos with plots, subtitles and scripts to generate answers. Tu et al. \cite{tu2014joint} also combine video and text data for question answering. Zhu et al. \cite{zhu2015uncovering} collect a dataset for "fill-in-the-blank" type of questions. Mazaheri et al. \cite{mazaheri2016video} also consider the fill-in-the-blank problem. Comparing with Image-QA, Video-QA is more troublesome because of the additional temporal dimension. The useful information is scattered in different frames. The temporal coherence must be well addressed to better understand the videos. Our proposed method focuses on the Multiple-Choice type of questions where the candidate answers are basically phrases. We collect the dataset by turning existing video description datasets automatically into Video-QA datasets which saves a lot of human labor. Moreover, we propose a model which can better utilize the temporal property of the videos and handle the answers in phrase form.

\section{Conclusion and Future Works}
We propose to collect a large-scale Video-QA dataset by automatically converting from the video description dataset. To tackle the Video-QA task, we propose two mechanisms: the re-watching and the re-reading mechanisms and then combine them into an effective forgettable-watcher model. In the future, we will improve the quality and increase the quantity of our dataset. We will also consider more QA types especially the open-ended QA problems.
\newpage
\bibliographystyle{named}
\bibliography{ijcai17}

\end{document}